\documentclass[runningheads]{llncs}
\usepackage{diagbox}
\usepackage{graphicx}
\usepackage{amsmath}
\usepackage{amssymb}
\usepackage{booktabs}
\usepackage{multirow}
\usepackage{url}
\usepackage{cuted}
\usepackage{placeins}
\usepackage{subcaption}
\usepackage{xcolor}
\usepackage[hidelinks]{hyperref}
\usepackage{wrapfig}
\usepackage{tabularx}
\usepackage{array}
\usepackage{caption}
\usepackage{float}
\begin{document}
\title{Asymmetric Cross-Modal Fine-Grained Visual Categorization: ACF-Net and the BirdPro Benchmark}
%
%

\author{%
Bohan Deng\inst{1} \and
Shuo Ye\inst{1}\textsuperscript{*} \and
Zitong Yu\inst{1}
}
\institute{%
Great Bay University\\
\email{bohandeng102647@gmail.com}\\
\email{shuoye.ke@gmail.com}\\
\email{zitong.yu@ieee.org}\\[0.3em]
{\small \textsuperscript{*}Corresponding author: Shuo Ye}
}
\titlerunning{ACF-Net and BirdPro}
\authorrunning{B. Deng et al.}
\maketitle            
\begin{abstract}
Audio-visual cross-modal Fine-Grained Visual Categorization~(FGVC) aims to identify fine-grained categories by jointly leveraging visual and auditory information. 
However, FGVC under asymmetric cross-modal scenarios has received limited attention, where paired video and audio are not strictly synchronized and may not even correspond to the same individual or moment. Such weak and ambiguous cross-modal correspondence poses substantial challenges to effective representation learning and modality alignment.
To address these issues, we propose ACF-Net, a novel optical flow-guided framework for asymmetric audio-visual fine-grained learning. ACF-Net consists of two key modules: Optical Flow-Guided Motion (OFGM) and Asymmetric Cross-Modal Adaptive Fusion (ACAF). OFGM captures motion-sensitive visual cues and suppresses irrelevant background interference, thereby enhancing discriminative dynamic representations in videos. ACAF estimates modality reliability under weakly matched audio-video pairs and performs uncertainty-aware adaptive fusion to improve category-level recognition robustness. To support research on asymmetric cross-modal FGVC, we further construct BirdPro, a new bird-oriented audio-visual benchmark, since existing datasets often lack large-scale category-level audio-video associations under non-strict temporal and instance correspondence. BirdPro contains 1,919 audio recordings and 11,965 videos covering 194 bird species. 
Extensive experiments show that ACF-Net achieves the best results compared with representative baseline methods, outperforming the strongest baselines by 2.97\% and 1.92\% in the fused and mismatched settings, respectively.

The code repository is available at \href{https://github.com/wxdbh/BirdPro-and-Fine-grained-Recognition}{wxdbh/BirdPro-and-Fine-grained-Recognition}.

\end{abstract}
\section{Introduction}
Fine-grained Visual Categorization~(FGVC) aims to distinguish highly similar subcategories or instances within the same semantic superclass ~\cite{peng2022balanced,11440120}. Its core objective is not merely to determine whether an object belongs to a general category, but to identify the specific fine-grained type of the object ~\cite{ye2023image}. As a result, fine-grained visual analysis relies heavily on weak discriminative cues, such as internal local structures, subtle textures, geometric proportions, and part composition relationships, to achieve precise differentiation ~\cite{fu2023multimodal,ye2022discriminative}. Related studies have been widely applied in various domains, including biological species identification, intelligent manufacturing, surveillance and security, cross-modal content retrieval, and medical image–assisted diagnosis ~\cite{lin2024samct,lin2023batformer}, playing an important role in improving perceptual accuracy under complex environments ~\cite{zhu2025emosym}. 

\begin{figure}[t]
  \begin{center}
    \includegraphics[width=0.89\textwidth]{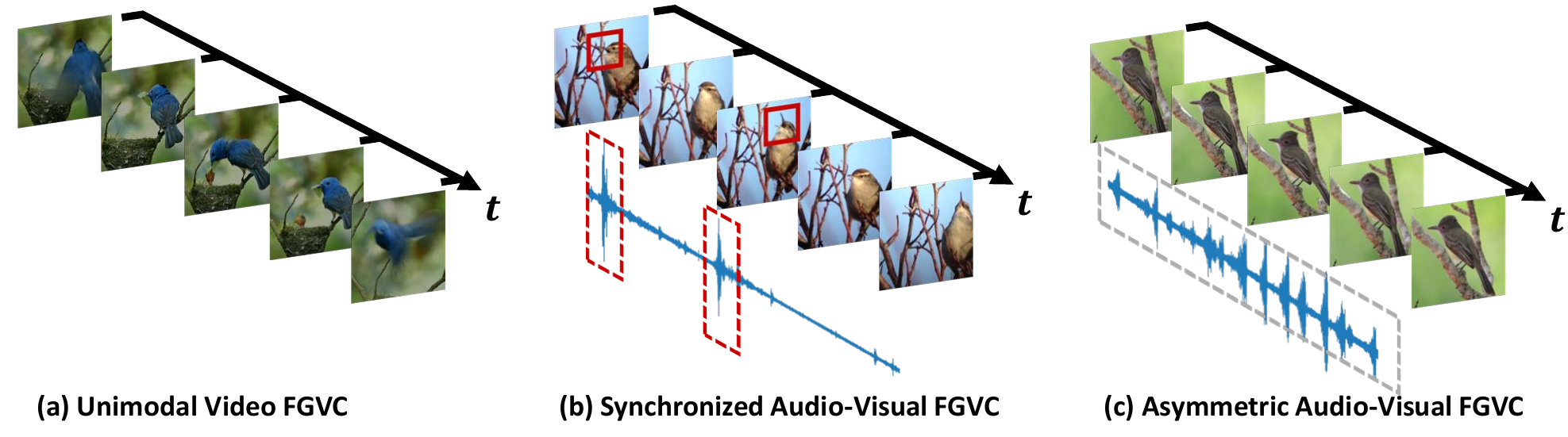}
  \end{center}
  \vspace{-0.3cm}
  \caption{Motivation of asymmetric audio-visual FGVC. 
(a) Unimodal video FGVC~\cite{he2022transfg} suffers from fragile visual cues under pose variation, occlusion, or background clutter. 
(b) Synchronized audio-visual FGVC~\cite{girdhar2023imagebind} leverages complementary acoustic cues under strict temporal and instance-level correspondence. 
(c) Asymmetric audio-visual FGVC only assumes category-level semantic consistency, where audio and video may differ in instance, time, or context.}
  \label{motivation}
\end{figure}

In real-world FGVC, subtle visual cues are easily degraded by pose variation, occlusion, background clutter, illumination changes, or limited image quality. Similar difficulties have also been observed in robust recognition and camouflaged/salient object perception, where category confusion, background camouflage, underwater degradation, and temporal instability motivate stronger discriminative and alignment mechanisms~\cite{ye2026ika2,hao2025simple,wu2026gbuucod,liu2026yuv20k}. Although video-based FGVC exploits appearance and motion, it is still limited by visual observability: discriminative parts may be occluded, key behaviors may be absent, and visually similar species can remain ambiguous. Additional modalities can provide complementary cues, but cross-modal heterogeneity in representation, information density, and noise distribution makes alignment and discriminative modeling challenging~\cite{girdhar2023imagebind,xie2025fg}.
Existing cross-modal FGVC methods often establish correspondences via vision-language adaptation, prompt learning, adapter tuning, or textual guidance, from category-level matching to attribute-, part-, or region-level alignment~\cite{girdhar2023imagebind}. However, they usually assume strict correspondence across modalities, such as the same instance, temporal segment, or synchronized context~\cite{peng2022balanced,fan2023pmr,fu2023multimodal}, as shown in Fig.~\ref{motivation}. This assumption is often violated in real-world audio-visual data, where audio and visual signals may share category-level semantics but differ in instance, time, or context. For example, bird calls can indicate species identity even when the observed bird is silent or the sound comes from another instance.
Therefore, cross-modal FGVC should exploit semantically related yet non-strictly aligned cues from asymmetric modalities. Inspired by expert cognition, where bird experts associate calls with appearance and behavior, we argue that learning such asymmetric semantic correlations is essential for robust cross-modal FGVC.

Therefore, we propose ACF-Net, an optical-flow-guided framework for asymmetric audio-visual cross-modal FGVC, to address the weak correspondence and heterogeneous characteristics between audio and visual modalities under asymmetric settings ~\cite{peng2022balanced,fan2023pmr,fu2023multimodal}. ACF-Net consists of two key modules: Optical Flow-Guided Motion (OFGM) and Asymmetric Cross-Modal Adaptive Fusion (ACAF). Specifically, OFGM enhances motion-sensitive visual representations by emphasizing category-relevant dynamic patterns while suppressing static background distractions. ACAF further performs uncertainty-aware adaptive fusion by estimating the reliability of each modality, allowing the model to rely more on the more informative modality under weakly associated audio-video pairs ~\cite{girdhar2023imagebind,xie2025fg}. This design improves the robustness of category-level recognition in asymmetric settings. Furthermore, we construct BirdPro, a new bird-oriented asymmetric cross-modal dataset, providing a new benchmark for asymmetric audio-visual FGVC.
The main contributions are summarized as follows:
\begin{itemize}
\item We revisit cross-modal FGVC under asymmetric audio-visual scenarios and show that semantically correlated discriminative cues can still be jointly learned even without strict instance-, time-, or context-level correspondence.
\item We propose an optical-flow-guided asymmetric audio-visual FGVC framework, which enhances motion-sensitive visual cues and adaptively aligns heterogeneous audio-visual representations for robust semantic fusion.
\item We construct BirdPro, a bird-oriented asymmetric audio-visual FGVC dataset, and conduct extensive experiments to demonstrate that the proposed method achieves state-of-the-art performance.
\end{itemize}

\section{Related Work}
{\bf Fine-grained Visual Categorization~(FGVC).}
FGVC aims to distinguish visually similar categories by capturing subtle discriminative cues~\cite{krause2015fine}. 
Early studies mainly relied on part-based modeling and region-level analysis, while recent convolutional and transformer-based methods improve fine-grained representation learning through multi-scale modeling, self-attention, and structured feature aggregation~\cite{wei2021fine,he2022transfg,fu2017look}. 
Related dense prediction and robust recognition studies further show that suppressing distractors and preserving weak target cues are crucial under category confusion, camouflage, and complex spatiotemporal backgrounds~\cite{ye2026ika2,hao2025simple,wu2026gbuucod,liu2026yuv20k}. 
In addition, class-incremental learning methods explore predictive prompting, embedding distillation, and task-oriented generation to preserve discriminative knowledge when new categories arrive~\cite{huang2026preprompt,huang2024etag}. 
Beyond image-based recognition, video-based FGVC further exploits temporal dynamics, and auditory fine-grained recognition shows that sound can provide complementary semantic cues for tasks such as bird species recognition and bioacoustic monitoring. 
However, most existing FGVC methods are developed under unimodal settings or assume reliable cross-modal correspondence. 
In real-world audio-visual scenarios, visual and auditory observations can be asymmetric, asynchronous, or partially missing, making it necessary to exploit complementary cues while suppressing noisy or weakly related modal information.

{\bf Cross-modal Alignment.}
Cross-modal alignment seeks to bridge heterogeneous modalities by learning semantically consistent representations~\cite{girdhar2023imagebind,xie2025fg}. 
Representative methods include shared embedding learning, contrastive learning, cross-modal attention, and collaborative encoding, which have shown effectiveness in cross-modal retrieval, multimodal classification, and vision-language understanding~\cite{hao2025simple}. 
For FGVC, however, alignment is more challenging because recognition often depends on sparse local cues, fine-grained attributes, and weakly salient patterns rather than global semantics alone. 
Recent studies therefore explore attribute- or token-level alignment to improve fine-grained discrimination~\cite{xie2025fg}. 
Nevertheless, most alignment methods still assume that informative cues from different modalities are synchronously observed and semantically consistent. 
This assumption is often violated in asymmetric audio-visual FGVC, where modalities may provide unequal, incomplete, or weakly correlated evidence~\cite{peng2022balanced,fan2023pmr,fu2023multimodal,subedar2019uncertainty}. 
Thus, the key challenge is not only reducing modality gaps, but also determining which cross-modal information is reliable and beneficial for fine-grained recognition~\cite{hao5089840distribution}.

{\bf Auxiliary Information-Guided Learning.}
Auxiliary information-guided learning improves representation learning by introducing additional priors or side information to guide the primary modality. 
Unlike conventional multimodal learning, auxiliary information is often used to constrain feature extraction or representation optimization, rather than directly serving as an independent prediction source. 
Typical guidance signals include semantic cues, such as textual descriptions or attribute labels~\cite{xie2025fg}; structural or motion cues, such as pose, segmentation masks, and optical flow~\cite{chen2025dynamic}; and weakly coupled cross-modal cues that provide auxiliary constraints without requiring strict alignment~\cite{chen2024causal}. 
Although these methods have shown promising results, many of them still rely on reliable correspondence between auxiliary information and the target modality~\cite{chen2025toward}. 
In asymmetric audio-visual FGVC, such correspondence can be weak, noisy, or temporally inconsistent. 
Therefore, how to exploit latent cross-modal correlations as effective guidance without enforcing strict correspondence remains an important problem.

\section{Proposed Method}
\begin{figure}[t]
	\begin{center}
		\includegraphics[width=0.9\textwidth]{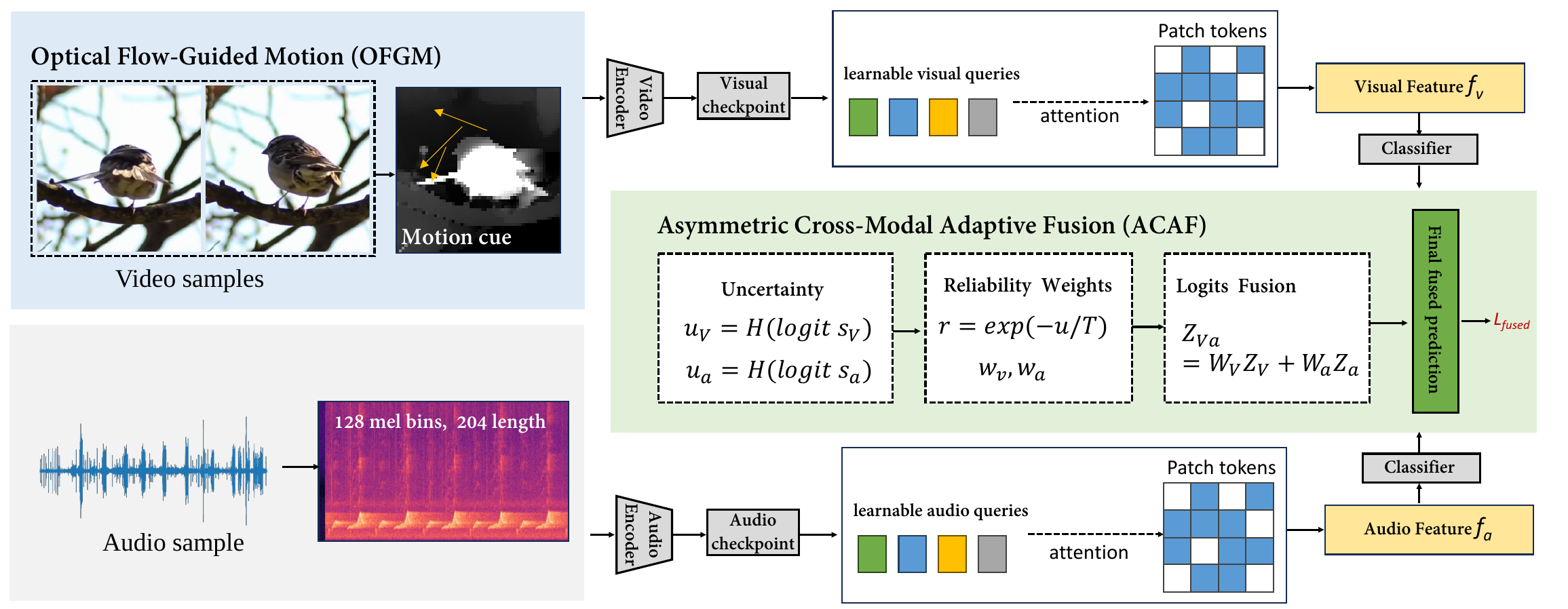}
	\end{center}
	\caption{Framework of ACF-Net. OFGM enhances RGB frames with optical-flow-guided motion cues, while ACAF adaptively fuses multimodal features for final classification.}
	\label{overview}
\end{figure}

The framework of ACF-Net is illustrated in Fig.~\ref{overview}.
Given an audio-visual bird sample, we denote the video clip and audio signal as $\mathcal{I}=\{I_t\}_{t=1}^{T}$ and $\mathcal{A}$, respectively, with a shared category label $y \in \{1,\ldots,K\}$. The two modalities are only category-aligned and may come from different temporal segments, recording conditions, or instances of the same bird species. Let $E_v(\cdot)$ and $E_a(\cdot)$ denote the visual and audio encoders, respectively~\cite{dosovitskiy2020image,xie2025fg,girdhar2023imagebind}. ACF-Net first applies OFGM to inject optical-flow-guided motion cues into RGB frames, producing a motion-aware visual representation $\tilde{\mathbf{z}}_v$. In parallel, the audio encoder extracts $\mathbf{z}_a$ from $\mathcal{A}$. Based on these representations, ACAF estimates modality reliability from prediction uncertainty and adaptively fuses visual and audio logits for final classification.

\subsection{Optical Flow-Guided Motion (OFGM)}
In asymmetric audio-visual fine-grained scenarios, visual clips provide dense temporal observations but often lack explicitly structured motion cues. Fine-grained bird recognition may rely on subtle dynamics, such as posture changes, beak movements, and wing flapping, which can be overwhelmed by static appearance or background variations. To address this issue, OFGM introduces optical flow as a motion prior to emphasize motion-sensitive visual regions~\cite{sun2024streamflow}.
Given an input video clip $\mathcal{I}=\{I_t\}_{t=1}^{T}$, we first compute optical flow maps between consecutive frames:
\begin{equation}
F_t = \operatorname{StreamFlow}(I_t, I_{t+1}), \quad t=1,\ldots,T-1.
\end{equation}
For each spatial position $(x,y)$, the optical flow vector is written as:
\begin{equation}
F_t(x,y) = \big(u_t(x,y), v_t(x,y)\big),
\end{equation}
where $u_t(x,y)$ and $v_t(x,y)$ denote horizontal and vertical displacement components, respectively. 
The motion intensity is computed by the flow magnitude:
\begin{equation}
S_t(x,y)=\sqrt{u_t(x,y)^2+v_t(x,y)^2}.
\end{equation}
We then normalize the magnitude map to obtain a motion-aware spatial mask:
\begin{equation}
M_t = \operatorname{Norm}(S_t),
\end{equation}
where $\operatorname{Norm}(\cdot)$ rescales the flow magnitude into $[0,1]$. 
The motion-enhanced RGB frame is generated by:
\begin{equation}
\tilde{I}_t = I_t \odot \left(\alpha + (1-\alpha)M_t\right),
\end{equation}
where $\odot$ denotes element-wise multiplication and $\alpha$ controls the minimum retained appearance intensity. 
When $M_t(x,y)$ is large, the corresponding visual region is preserved or emphasized; when $M_t(x,y)$ is small, the region is weakened but not completely removed. 
This design preserves appearance information while injecting motion-sensitive guidance.

The motion-enhanced video clip is denoted as:
\begin{equation}
\tilde{\mathcal{I}} = \{\tilde{I}_t\}_{t=1}^{T}.
\end{equation}
For the last frame, we reuse the preceding motion mask, i.e., $M_T=M_{T-1}$. The visual representation is then extracted by the visual encoder:
\begin{equation}
\tilde{\mathbf{z}}_v = E_v(\tilde{\mathcal{I}}).
\end{equation}

Unlike direct RGB-flow feature concatenation, this mask-based formulation uses optical flow only as a spatial prior, avoiding an additional motion encoder while guiding the visual encoder toward fine-grained foreground dynamics.

\subsection{Asymmetric Cross-Modal Adaptive Fusion (ACAF)}
Fixed-weight fusion is often suboptimal for asymmetric audio-visual fine-grained recognition~\cite{peng2022balanced,fan2023pmr,fu2023multimodal}, since modality reliability varies across samples. Audio signals may contain clear species-specific vocal patterns or be noisy and weakly related, while visual observations may suffer from occlusion, background clutter, or limited motion. Therefore, ACAF performs sample-adaptive fusion according to modality uncertainty.

Given the motion-enhanced visual representation $\tilde{\mathbf{z}}_v$ and the audio representation $\mathbf{z}_a$, we first obtain modality-specific logits:
\begin{equation}
\mathbf{o}_v = C_v(\tilde{\mathbf{z}}_v), \quad
\mathbf{o}_a = C_a(\mathbf{z}_a),
\end{equation}
where $C_v(\cdot)$ and $C_a(\cdot)$ denote the visual and audio classifiers, respectively.

We estimate modality uncertainty from the prediction distributions:
\begin{equation}
\mathbf{p}_v = \operatorname{Softmax}(\mathbf{o}_v), \quad
\mathbf{p}_a = \operatorname{Softmax}(\mathbf{o}_a).
\end{equation}
The entropy-based uncertainty of modality $m \in \{v,a\}$ is computed as:
\begin{equation}
u_m = -\sum_{k=1}^{K} p_m^k \log p_m^k.
\end{equation}
A lower entropy indicates a more confident and reliable modality, while a higher entropy indicates a more uncertain prediction.

We convert uncertainty into reliability scores:
\begin{equation}
\rho_v = \exp(-u_v / \tau), \quad
\rho_a = \beta \exp(-u_a / \tau),
\end{equation}
where $\tau$ is a temperature parameter and $\beta>0$ is a learnable audio scaling factor.
The adaptive fusion weights are obtained by normalization:
\begin{equation}
w_v = \frac{\rho_v}{\rho_v+\rho_a}, \quad
w_a = \frac{\rho_a}{\rho_v+\rho_a}.
\end{equation}

The fused logits are then computed as:
\begin{equation}
\mathbf{o}_{va} = w_v \mathbf{o}_v + w_a \mathbf{o}_a.
\end{equation}
This uncertainty-aware design allows ACAF to assign higher weights to the more reliable modality for each sample. 
The training objective combines fused and modality-specific classification losses:
\begin{equation}
\mathcal{L}
=
\mathcal{L}_{\mathrm{fused}}
+
\lambda_v \mathcal{L}_{v}
+
\lambda_a \mathcal{L}_{a}
+
\lambda_p \mathcal{L}_{\mathrm{proto}}^{a}
+
\lambda_d \mathcal{L}_{\mathrm{dec}} .
\end{equation}
Here, $\mathcal{L}_{\mathrm{fused}}$, $\mathcal{L}_{v}$, and $\mathcal{L}_{a}$
are cross-entropy losses for fused, visual, and audio predictions, respectively.
$\mathcal{L}_{\mathrm{proto}}^{a}$ denotes the audio prototype regularization loss,
and $\mathcal{L}_{\mathrm{dec}}$ denotes the concept decoupling regularization.
The coefficients $\lambda_v$, $\lambda_a$, $\lambda_p$, and $\lambda_d$
control the contributions of the visual classification, audio classification,
audio prototype, and decoupling losses, respectively.

\begin{equation}
\mathcal{L}_{\mathrm{proto}}^{a}
=
\operatorname{CE}
\left(
\frac{
\operatorname{Norm}(\mathbf{z}_a)
\operatorname{Norm}(\mathbf{P}_a)^{\top}
}{\tau_p},
y
\right),
\end{equation}
where $\tau_p$ is the prototype temperature and
$\mathbf{P}_a \in \mathbb{R}^{K \times d}$ denotes the audio prototype matrix.

\begin{equation}
\mathbf{P}_{a}^{(k)}
\leftarrow
\mu \mathbf{P}_{a}^{(k)}
+
(1-\mu)
\operatorname{Norm}
\left(
\frac{1}{|\mathcal{B}_k|}
\sum_{i \in \mathcal{B}_k}
\mathbf{z}_{a,i}
\right),
\end{equation}
where $\mathbf{P}_{a}^{(k)}$ is the audio prototype of class $k$,
$\mu$ is the EMA momentum, and $\mathcal{B}_k$ denotes the set of samples
belonging to class $k$ in the current mini-batch.

\begin{equation}
\mathcal{L}_{\mathrm{dec}}^{m}
=
\frac{1}{B}
\sum_{i=1}^{B}
\frac{1}{Q_m(Q_m-1)}
\sum_{\substack{r,s=1 \\ r \neq s}}^{Q_m}
\left(
\hat{\mathbf{c}}_{i,r}^{m\top}
\hat{\mathbf{c}}_{i,s}^{m}
\right)^2,
\quad m \in \{v,a\}.
\end{equation}
Here, $B$ denotes the mini-batch size, $Q_m$ is the number of concept
queries for modality $m$, and $\hat{\mathbf{c}}_{i,r}^{m}$ denotes the
L2-normalized feature of the $r$-th concept query for the $i$-th sample in
modality $m$.

\begin{equation}
\mathcal{L}_{\mathrm{dec}}
=
\mathcal{L}_{\mathrm{dec}}^{v}
+
\mathcal{L}_{\mathrm{dec}}^{a}.
\end{equation}

\section{Experiments}
\subsection{BirdPro Dataset}
We adopt the species-level semantic annotations provided by the CUB dataset~\cite{wah2011caltech} as a unified label space, and use the corresponding scientific names as query keywords to systematically retrieve videos and audio data from publicly available online resources. Based on this process, we construct a multi-modal fine-grained bird dataset, ensuring semantic consistency and comparability across different modalities.
The dataset statistics are shown in Table~\ref{dataset_statistics}.
\begin{wraptable}{r}{0.38\textwidth}
\vspace{-10pt}
\centering
\small
\setlength{\tabcolsep}{4pt}
\renewcommand{\arraystretch}{1.15}
\caption{Statistics of BirdPro. ``M'' denotes the modality type.}
\label{dataset_statistics}
\vspace{2pt}
\resizebox{\linewidth}{!}{%
\begin{tabular}{c|c|c|c}
\hline
\textbf{M} & \textbf{Classes} & \textbf{Train} & \textbf{Test} \\
\hline
Video  & 194 & 10,054 & 1,911 \\
Audio  & 194 & 1,531  & 388   \\
Pair   & 194 & 6,022  & 470   \\
\hline
\end{tabular}}
\vspace{-4pt}
\end{wraptable}
During data collection, candidate videos are first manually screened to ensure that the target species is visually identifiable and occupies a dominant region in a sufficient number of frames. From the selected videos, representative key frames are extracted to cover diverse viewpoints, poses, and environmental conditions, thereby preserving fine-grained visual details that are critical for species discrimination. In parallel, audio samples corresponding to the same species are collected from independent sources, with preference given to segments featuring low background noise and clear acoustic structures.
The resulting dataset provides a more challenging and realistic benchmark for studying cross-modal FGVC and robust representation learning under complex environmental conditions. 

\subsection{Implementation Details}
Our framework adopts a two-stream audio-visual architecture. 
For the visual branch, we use ViT-B/16 with an input resolution of $448$, and each video is represented by 6 sampled frames. 
For the audio branch, we follow the ImageBind~\cite{girdhar2023imagebind}-style audio representation and use log-mel spectrogram inputs with a target length of 204. 
To enhance motion-sensitive visual representation, StreamFlow ~\cite{sun2024streamflow} is used to generate compact $28 \times 28$ optical flow maps as motion priors, which suppress background-dominant motion while preserving efficient storage and computation.
For multimodal fusion, the feature dimension is set to 512, and the dropout rate is set to 0.1. 
During fine-tuning, the last 6 blocks of the audio encoder and the last 2 layers of the visual encoder are unfrozen, while the remaining parameters are kept fixed to preserve pretrained representations. 
Unless otherwise specified, the same configuration is used for all ablation and comparison experiments.
We evaluate the proposed method under both unimodal and multimodal settings. 

\begin{figure}[hbpt]
\centering
\includegraphics[width=\textwidth]{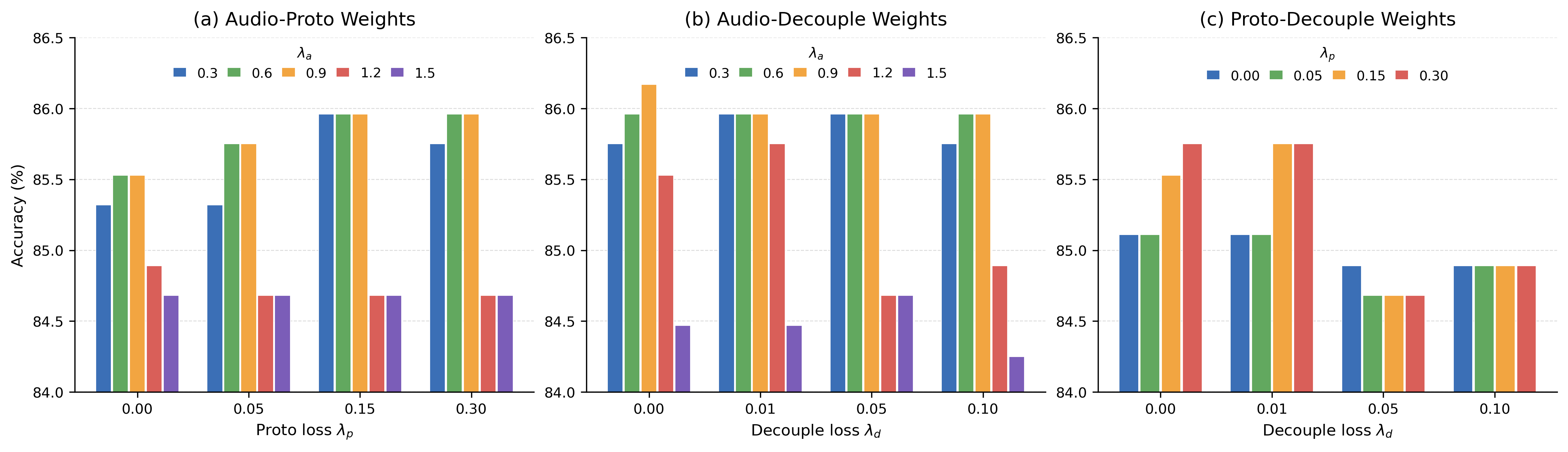}
\caption{Pairwise sensitivity analysis of loss weights. 
}
\label{fig:pairwise_sensitivity}
\end{figure}

To further evaluate robustness under asymmetric multimodal correspondence, we conduct an audio-visual mismatch experiment. 
Specifically, we apply a class-level derangement to the training split, where audio samples from class $c$ are reassigned to another class $\pi(c)$ with $\pi(c)\neq c$, while the visual samples remain unchanged. 
The validation split is kept clean. 
This setting tests whether each fusion strategy can tolerate corrupted audio-visual correspondence during training and still perform reliable audio-visual recognition during evaluation.

\subsection{Model Configuration}
\noindent\textbf{Loss-weight sensitivity.} As shown in Fig.~\ref{fig:pairwise_sensitivity}, a moderate weight for the audio prototype loss generally improves performance, suggesting that prototype regularization enhances the discriminability of audio representations. 
For the decoupling loss, overly large weights tend to degrade accuracy, implying that excessive concept separation may weaken useful shared information. 
The best performance is achieved with a balanced combination of prototype and decoupling losses, demonstrating that these regularization terms are complementary when properly weighted.

\noindent\textbf{Temporal sampling and concept queries.} We further investigate the effects of the number of sampled visual frames, audio clips, and modality-specific concept queries.
\begin{table}[t]
\centering
\small
\caption{Sensitivity analysis of key hyperparameters.}
\label{tab:sensitivity}
\begin{subtable}[t]{0.47\linewidth}
\centering
\caption{Frames and audio clips}
\label{tab:sens_frames_clips}
\setlength{\tabcolsep}{4pt}
\begin{tabular}{l c c c}
\toprule
\diagbox{\textbf{Frames}}{\textbf{Clips}}
 & \textbf{10} & \textbf{20} & \textbf{30} \\
\midrule
2 & 83.40 & 83.19 & 84.25 \\
4 & 85.75 & 84.68 & 86.17 \\
6 & 85.75 & 85.53 & 87.23 \\
8 & 86.81 & 84.47 & 86.38 \\
\bottomrule
\end{tabular}
\end{subtable}
\hfill
\begin{subtable}[t]{0.47\linewidth}
\centering
\caption{Impact of query numbers~($v_N$, $a_N$)}
\label{tab:sens_vce_audio_vce}
\setlength{\tabcolsep}{4pt}
\begin{tabular}{l c c c c}
\toprule
\diagbox{\textbf{$v_N$}}{\textbf{$a_N$}}
\textbf{} & \textbf{2} & \textbf{4} & \textbf{6} & \textbf{8} \\
\midrule
3 & 87.02 & 84.04 & 84.68 & 87.02 \\
5 & 85.75 & 84.68 & 86.38 & 84.89 \\
7 & 84.68 & 85.11 & 85.11 & 84.68 \\
9 & 84.89 & 85.11 & 85.11 & 86.60 \\
\bottomrule
\end{tabular}
\end{subtable}
\end{table}
As shown in Table~\ref{tab:sens_frames_clips}, using too few visual frames, e.g., 2 frames, leads to inferior performance, since limited visual observations are insufficient to capture fine-grained motion and pose variations. 
Increasing the number of frames generally improves the recognition accuracy. However, further increasing the visual frames does not consistently bring additional gains, suggesting that excessive frames may introduce redundant or less discriminative information. 
Therefore, we adopt 6 visual frames and 30 audio clips as the default temporal sampling configuration.
For the number of visual and audio queries, the results in Table~\ref{tab:sensitivity} show that simply increasing the number of queries does not always improve performance. A small number of visual queries can already capture discriminative fine-grained visual concepts, while excessive queries may introduce redundant concepts and increase optimization difficulty. For the audio branch, different query numbers lead to performance fluctuations, indicating that audio concept modeling is sensitive to the balance between representation capacity and noise suppression. Based on these observations, we choose a moderate query configuration that achieves competitive performance while avoiding unnecessary redundancy.

\subsection{Comparison Results}
Table~\ref{tab:main_results} reports the comparison results across different modality settings.
\begin{table}[t]
\centering
\caption{Comparison results. $^{*}$ denotes our implemented baseline variant based on the cited method. The best results are highlighted in bold.}
\label{tab:main_results}
\vspace{4pt}
\scriptsize
\setlength{\tabcolsep}{4pt}
\renewcommand{\arraystretch}{0.90}
\begin{tabular}{lcccc|c}
\toprule
\textbf{Method} & \textbf{M} & \textbf{Audio} & \textbf{Video} & \textbf{Fusion} & \textbf{Mismatch} \\
\midrule
FG-CLIP~\cite{xie2025fg} & V-only & -- & 70.30 & -- & -- \\
ACF (ours) & V-only & -- & 71.06 & -- & -- \\
\midrule
ImageBind~\cite{girdhar2023imagebind} & A-only & 47.42 & -- & -- & -- \\
ACF (ours) & A-only & 57.45 & -- & -- & -- \\
\midrule
CoC$^{*}$~\cite{kittler1998combining} & A + V & 45.74 & 71.06 & 80.43 & 70.21 \\
MDL$^{*}$~\cite{ngiam2011multimodal} & A + V & 47.73 & 71.92 & 73.40 & 70.43 \\
WeC$^{*}$~\cite{kittler1998combining} & A + V & 42.34 & 71.06 & 82.34 & 69.79 \\
UAF$^{*}$~\cite{subedar2019uncertainty} & A + V & 47.87 & 71.91 & 83.19 & 72.13 \\
UDF$^{*}$~\cite{subedar2019uncertainty} & A + V & 49.36 & 70.21 & 84.26 & 72.34 \\
DMI$^{*}$~\cite{liu2022dense} & A + V & 41.70 & 70.64 & 75.32 & 70.00 \\
CoS$^{*}$~\cite{subedar2019uncertainty} & A + V & 42.77 & 71.49 & 79.15 & 72.34 \\
GMU$^{*}$~\cite{arevalo2017gated} & A + V & 49.15 & 71.91 & 75.11 & 70.64 \\
CBP$^{*}$~\cite{fukui2016multimodal} & A + V & 42.98 & 72.34 & 77.66 & 52.55 \\
ACF-Net (ours) & A + V & \textbf{58.30} & \textbf{72.77} & \textbf{87.23} & \textbf{74.26} \\
\bottomrule
\end{tabular}
\vspace{-6pt}
\end{table}

\noindent\textbf{Unimodal results.} ACF-Net improves the video-only accuracy from 70.30\% to 71.06\% compared with FG-CLIP ~\cite{xie2025fg}, and improves the audio-only accuracy from 47.42\% to 57.45\% compared with ImageBind ~\cite{girdhar2023imagebind}. These results indicate that the proposed framework enhances both visual and audio representations, with a more substantial gain on the weaker audio modality.

\noindent\textbf{Multimodal results.} Under this setting, direct fusion variants such as feature concatenation, cross-modal attention, gated fusion, and bilinear fusion do not consistently improve performance. This suggests that symmetric fusion can introduce unreliable modality interactions in asymmetric audio-visual FGVC. In contrast, uncertainty-aware methods achieve stronger results, confirming the importance of estimating modality reliability. ACF achieves the best fusion accuracy of 87.23\%, outperforming the strongest baseline by 2.97\%, which demonstrates its ability to exploit complementary audio-visual information while reducing the influence of unreliable audio cues.

\noindent\textbf{Mismatch robustness.} To further evaluate robustness under unreliable cross-modal correspondence, we conduct an audio-visual mismatch experiment. Specifically, audio samples are reassigned to semantically unrelated visual classes during training, while the validation split remains clean. This setting simulates asymmetric multimodal scenarios where the audio modality may be weak, noisy, or incorrectly paired with the visual modality.
As can be seen, conventional fusion strategies suffer from clear performance degradation under mismatched audio-visual supervision. In particular, Compact Bilinear Pooling drops to 52.55\%, indicating that strong multiplicative interaction is sensitive to corrupted cross-modal correspondence. Uncertainty-aware methods are more robust because they reduce the influence of unreliable modality information. ACF obtains the best accuracy of 74.26\%, outperforming the strongest baseline by 1.92\%, demonstrating its robustness under asymmetric multimodal conditions.

\subsection{Ablation Study}
To evaluate the effectiveness of the proposed method, we conduct comprehensive ablation studies on our benchmark dataset. \textbf{Baseline.} We first report the performance of a unimodal visual recognition framework as the \emph{baseline}, which serves as a reference for assessing the contribution of each proposed component under the asymmetric cross-modal fine-grained setting.
\textbf{OFGM.} We then introduce the setting \emph{w/ OFGM} to evaluate the overall performance gain brought by the \emph{Optical Flow-Guided Fine-grained Motion Modeling} module.
\textbf{ACAF.} Next, we adopt the setting \emph{w/ ACAF} to assess the impact of the \emph{Asymmetric Cross-Modal Adaptive Fusion} module.
\textbf{Full model.} Finally, we combine both modules in the full model to examine whether OFGM and ACAF provide complementary benefits.
The experimental results are summarized in Table~\ref{tab:ablation}.

\begin{wraptable}{r}{0.46\textwidth}
\vspace{-10pt}
\centering
\scriptsize
\caption{Ablation study results.}
\label{tab:ablation}
\setlength{\tabcolsep}{3pt}
\renewcommand{\arraystretch}{0.95}
\resizebox{\linewidth}{!}{
\begin{tabular}{l c c c}
\toprule
\textbf{Method} & \textbf{Audio} & \textbf{Video} & \textbf{Fusion} \\
\midrule
Baseline & 45.96 & 71.06 & 82.98 \\
w/ OFGM   & 46.38 & 71.92 & 81.64 \\
w/ ACAF  & 57.66 & 72.77 & 84.89 \\
Ours     & \textbf{58.30} & \textbf{72.77} & \textbf{87.23} \\
\bottomrule
\end{tabular}
}
\vspace{-12pt}
\end{wraptable}
\noindent\textbf{Effect of OFGM.} As can be seen, the baseline model achieves 82.98\% fusion accuracy, while introducing OFGM improves the video accuracy from 71.06\% to 71.92\%. 
This result indicates that optical-flow-guided motion cues can enhance the visual branch by capturing fine-grained dynamic patterns that are difficult to identify from RGB frames alone. 
However, the fusion accuracy slightly decreases when only OFGM is introduced, suggesting that improving the visual representation alone does not necessarily lead to better multimodal fusion under asymmetric audio-visual conditions.

\noindent\textbf{Effect of ACAF.} When ACAF is adopted, the audio accuracy is significantly improved from 45.96\% to 57.66\%, and the fusion accuracy increases from 82.98\% to 84.89\%. 
This demonstrates that ACAF effectively alleviates the imbalance between the stronger visual modality and the weaker audio modality by adaptively estimating modality reliability. 

\noindent\textbf{Full model.} Compared with the baseline, the full model achieves the best performance, with 58.30\% audio accuracy, 72.77\% video accuracy, and 87.23\% fusion accuracy. 
These results verify that OFGM and ACAF provide complementary benefits: OFGM strengthens fine-grained motion-sensitive visual representation, while ACAF enables robust asymmetric cross-modal fusion. 
Their combination leads to the most effective multimodal representation and achieves the best overall recognition performance.

\subsection{Visualization}
To further analyze the effect of the proposed modules, we provide qualitative visualizations from both feature-space and spatial perspectives.

\noindent\textbf{t-SNE visualization~\cite{van2008visualizing}.} We visualize the feature distributions of baseline and ACF-Net representations for the video and audio modalities.
\begin{figure}[hbpt]
	\centering
	\includegraphics[width=\textwidth]{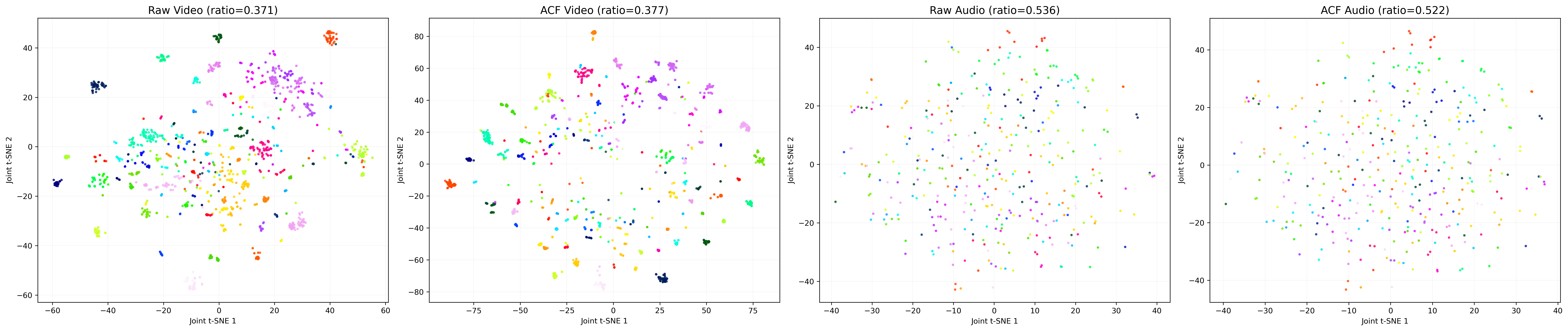}
	\caption{t-SNE visualization of raw and ACF-enhanced video and audio features. ACF improves the intra-class compactness of audio features while preserving the discriminative structure of video features.}
	\label{fig:tsne_all}
\end{figure}
As shown in Figure~\ref{fig:tsne_all}, the video modality exhibits clearer discriminative structures than the audio modality, indicating that visual inputs provide strong appearance-based cues for FGVC. After applying ACF-Net, video features form more compact and better-separated clusters, suggesting that the proposed framework further enhances visual discriminability by refining motion-aware representations. In contrast, raw audio features are more scattered and highly entangled across categories, reflecting the difficulty of learning discriminative representations from noisy and variable acoustic signals. Compared with raw audio features, ACF-Net enhanced audio features show improved intra-class compactness and a more coherent feature distribution. 

\begin{figure}[t]
\centering
\includegraphics[width=0.8\textwidth]{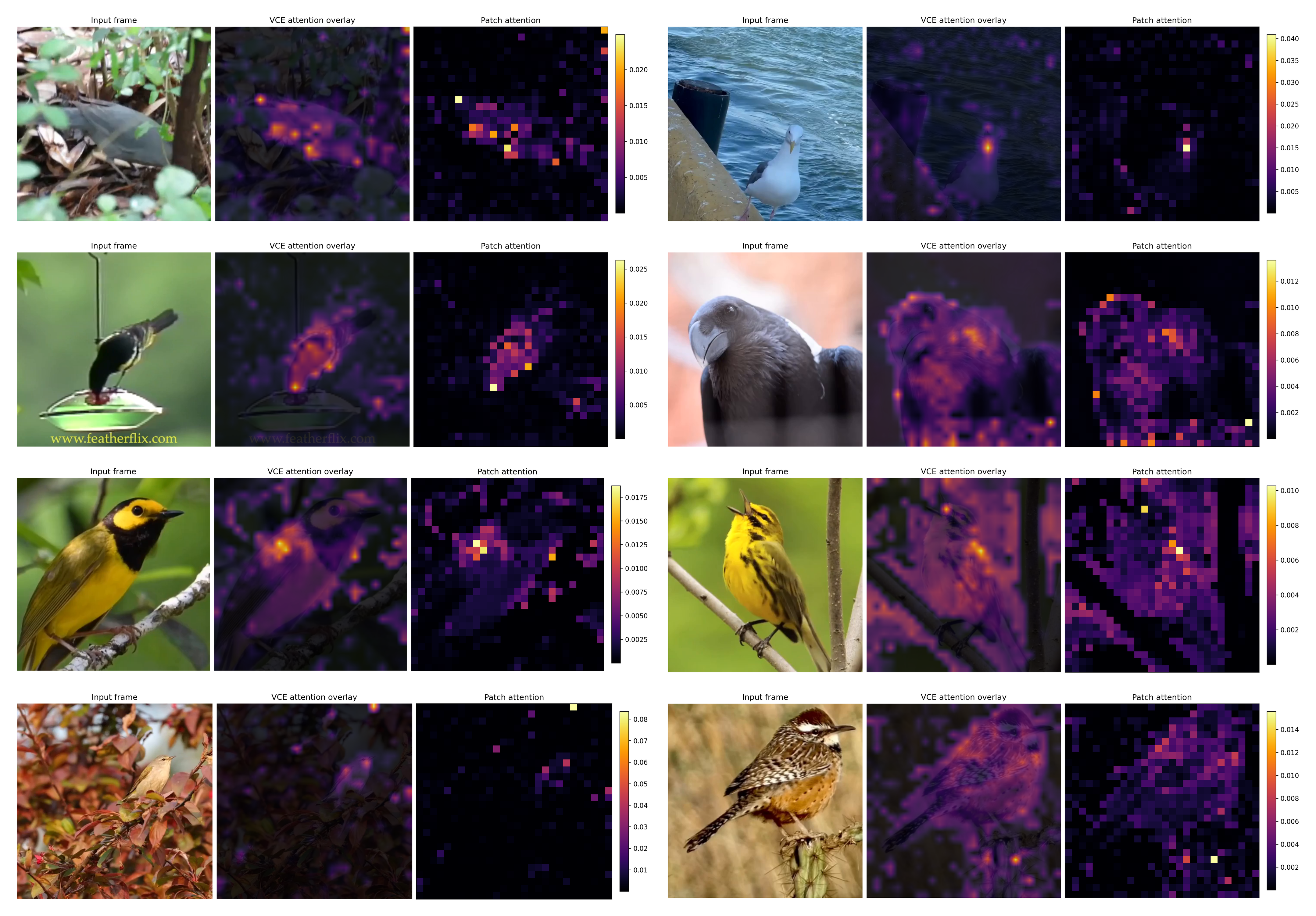}
\caption{Attention visualization of the proposed framework. For each example, we show the input frame, attention overlay, and patch-level attention map.}
\label{fig:vision_attention}
\end{figure}

\noindent\textbf{Attention visualization.} We further visualize the attention responses of the proposed method in Figure~\ref{fig:vision_attention} to examine whether the model focuses on semantically meaningful image regions.
The visualization shows that the proposed framework highlights the main object regions, especially the head, neck, and body, while suppressing irrelevant background areas. The corresponding patch-level attention maps also exhibit stronger responses around semantically informative patches. This suggests that our method guides the model toward discriminative visual regions and provides more interpretable evidence for the learned visual representation. These qualitative results are consistent with the quantitative improvements reported in previous sections.

\section{Conclusion}
In this paper, we investigate cross-modal FGVC under asymmetric audio-visual scenarios, where strict instance- or time-level correspondence is often unavailable.  We propose ACF-Net, an optical flow-guided framework that enhances motion-sensitive visual cues and performs reliability-aware adaptive fusion under weakly matched audio-video pairs.  We also construct the BirdPro dataset, a bird-oriented benchmark for asymmetric audio-visual FGVC.  Extensive experiments demonstrate the effectiveness and robustness of ACF-Net, showing its potential for fine-grained recognition in realistic non-ideal cross-modal settings.

%
%
%
%
\bibliographystyle{splncs04}
\bibliography{references}
\end{document}